\documentclass{ieeeaccess}
\usepackage{cite}
\usepackage{amsmath,amssymb,amsfonts}
\usepackage{algorithmic}
\usepackage{graphicx}
\usepackage{textcomp}
\usepackage{blindtext}
\usepackage[T1]{fontenc}
\usepackage{epstopdf}
\usepackage{multirow}
\usepackage{booktabs}
\usepackage{ragged2e}
\usepackage[inline]{enumitem}

\newcommand{\change}[1]{#1}

\def\BibTeX{{\rm B\kern-.05em{\sc i\kern-.025em b}\kern-.08em
		T\kern-.1667em\lower.7ex\hbox{E}\kern-.125emX}}
\begin{document}
	\history{Date of first submission Feburary 09, 2018. Date of resubmission May 09, 2018. Date of acceptance June 08, 2018.}
	\doi{10.1109/ACCESS.2018.2846543}
	
	\title{Slice as an Evolutionary Service: Genetic Optimization for Inter-Slice Resource Management in 5G Networks}
	\author{\uppercase{Bin Han}, \IEEEmembership{Member, IEEE},
		\uppercase{Ji Lianghai}, \IEEEmembership{Student Member, IEEE},\\
		\uppercase{and Hans D. Schotten}, \IEEEmembership{Member, IEEE}}
	\address{Institute of Wireless Communications, Technische Universit\"at Kaiserslautern, 67655 Kaiserslautern, Germany (e-mail: \{binhan,ji,schotten\}@eit.uni-kl.de)}
	\tfootnote{\justify This work has been supported by the European Union Horizon-2020 project 5G-MoNArch (grant agreement number 761445) and the Network for the Promotion of Young Scientists (TU-Nachwuchsring) at Technische Universit\"at Kaiserslautern (individual funding). The authors would like to acknowledge the contributions of their colleagues. This information reflects the consortium's view, but the consortium is not liable for any use that may be made of any of the information contained therein.}
	
	\markboth
	{B. Han \headeretal: Slice as an Evolutionary Service}
	{Genetic Optimization for Inter-Slice Resource Management in 5G Networks}
	
	\corresp{Corresponding author: Bin Han (e-mail: binhan@eit.uni-kl.de).}
	
	\begin{abstract}
		In the context of Fifth Generation (5G) mobile networks, the concept of \change{``Slice as a Service'' (SlaaS)} promotes mobile network operators to flexibly share infrastructures with mobile service providers and stakeholders. However, it also challenges with an emerging demand for efficient online algorithms to optimize the request-and-decision-based inter-slice resource management strategy. Based on genetic algorithms, this paper presents a novel online optimizer that efficiently approaches towards the ideal slicing strategy with maximized long-term network utility. The proposed method \change{encodes slicing strategies into binary sequences to cope with the request-and-decision mechanism. It} requires no a priori knowledge about the traffic/utility models, \change{and therefore supports heterogeneous slices,} while providing solid effectiveness, good robustness against non-stationary service scenarios, and high scalability.
	\end{abstract}
	
	
	\titlepgskip=-15pt
	
	\maketitle
	
	\section{Introduction}\label{sec:introduction}
	\PARstart{N}{etwork} slicing was proposed by the \textit{Next Generation Mobile Networks (NGMN) Alliance} \cite{alliance20155g}. Since then, it has become one of the hottest topics in the filed of future 5\textsuperscript{th} Generation (5G) mobile communication networks. Generally, the concept of network slicing can be understood as creating and maintaining multiple independent logical networks on a common physical infrastructure \change{platform}, every slice operates a separate business service with certain Quality of Service (QoS) requirements. Enabled and supported by the emerging technologies of software defined networks (SDN) and network function virtualization (NFV), network slicing exhibits great potentials -- as \change{indicated} in \cite{rost2017network} -- not only in supporting specialized applications with extreme performance requirements, but also in benefiting the mobile network operators (MNOs) with increased revenue.
	A sliced mobile network manages its infrastructure and virtual resources in independent scalable slices, each slice runs a homogeneous service with \change{a specific} business model. Thus, an MNO can dynamically and flexibly create, terminate and scale its slices to optimize the resource utilization.

	In a previous paper \cite{han2017modeling}, we have proposed a profit optimization model for sliced mobile networks that applies on the traditional business mode: the MNOs with network resources implement the slices and provide all network services directly to their end-users. In this case, a MNO \change{is fully aware of} a priori knowledge about the service demands and the cost/revenue models of every slice. It is able to scale the slices in real time according to their utility efficiencies \change{(which can be flexibly defined as such like some QoS or the revenue rate)}, in order to achieve the maximal overall network utility under the resource constraints. This is a classical multi-objective optimization problem (MOOP), in which the main challenge is to solve the optimal resource allocation, or at least to find a satisfactory solution, with affordable computing efforts.
	
	Unfortunately, this model does not apply on the slices operated by tenants such as mobile virtual network operators (MVNOs), which are considered to play an important role in 5G networks \cite{rost2016mobile}. Tenants are third-parties that provide services without owning any network infrastructure, some instances are utility/automotive companies and over-the-top service providers such as \textit{YouTube}\textsuperscript{\textregistered}. To provide connection services, they have to be granted by MNOs with network resources, including radio/infrastructure resources and virtualized resource blocks, i.e. computation resources. In legacy networks, every tenant makes its contractual agreement with the MNO(s), to pay a fixed and coarsely estimated annual/monthly fee for these resource sharing concepts. In the context of network slicing, in contrast, the resources are first bundled into slices before granted to tenants upon demand. Depending on the slice type, different slices have various utility efficiencies and periodical payments. For example, with the same amount of resource, slices for \change{mobile} broadband (MBB) services\change{require high} average user throughput, while slices for massive Machine-Type Communication (mMTC) services focus more to simultaneously serve more low-traffic devices \cite{5g2016view}. Even with the same service type, elastic slices can be defined to guarantee an average QoS level for a lower payment, while inelastic slices provide guaranteed minimal QoS level for a higher payment \cite{bega2017optimising}. This approach, usually known as \change{``Slice as a Service'' (SlaaS)} \cite{sciancalepore2017slice}, improves the sharing efficiency and the resource utilization rate. However, as such slices are operated by tenants, and their scales shall be formulated and protected by contractual agreements, a new agreement between the MNO and tenant may therefore be essential to flexibly rescale or terminate a slice at arbitrary time, which leads to extra operations expenditure (OPEX). As an efficient alternative, the MNO can offer resources to implement slices of different types, and macroscopically optimize its resource allocation by choosing if to accept or decline every request from tenant for slice creation. On the other hand, as every slice is logically isolated from the others, a tenant has no access to slices operated by other tenants, but only the administration over its own resources by requesting new slices or terminating active slices of its own. In this case, neither the MNO nor the tenants can jointly optimize all slices in a fully dynamic approach, which \change{disables most classical techniques of resource allocation and }proposes a new challenge of network resource management.
	
	First efforts have been made recently on this emerging topic. On the one hand, focusing on how the tenants \change{adjust} slice parameters to reduce cost while maintaining the quality of service, a game theory model has been proposed in \cite{caballero2017network}. On the other hand, taking the MNO's point of view, the authors of \cite{bega2017optimising} have proposed to optimize the slicing strategy in order to maximize the overall revenue. In this paper, we focus on the latter problem, and propose a novel online genetic approach of slicing strategy optimization. \change{Compared to existing methods, our proposed approach encodes every feasible slicing strategy into an individual binary sequence, so that it copes with the binary-decision based inter-slice control mechanism. Furthermore, it requires no pre-knowledge about the utility model, and therefore allows heterogeneous utility functions for different slice types, in order to better support the coexistence of heterogeneous slices with highly various QoS requirements in 5G networks.}
	
	The rest part of this paper is organized as follows: In Sec. \ref{sec:model}, we setup the system model to describe the business process of \change{SlaaS}. Then in Sec. \ref{sec:review} we review the existing methods \change{of resource allocation, especially the Q-Learning method in \cite{bega2017optimising}, and discuss about their limits}. Afterwards, Sec. \ref{sec:ga} briefly introduces genetic algorithms to help readers understand our proposed method which we present in Sec. \ref{sec:method}. Subsequently, we evaluate the performance of our approach through numerical simulations in Sec. \ref{sec:evaluation}, before Sec. \ref{sec:conclusion} closes the paper with conclusions and some outlooks.
	
	\section{System Model}\label{sec:model}
	\Figure[htbp!](topskip=0pt, botskip=0pt, midskip=0pt)[width=.6\textwidth]{System_Model.eps}
	{\change{System model of inter-slice resource management based on tenant requests and binary decisions. $M=N=2$ taken for the illustration.\label{fig:system_model}}}
	\subsection{Space of Resource Feasibility}
	Consider a MNO with $M$ different types of resources to support the maintenance of up to $N$ different types of slices. The resource pool can be therefore described with a $M$-dimensional vector $\mathbf{r}=[r_1,r_2,\dots,r_M]^\mathrm{T}$. Every slice type $n\in\{1,\dots,N\}$ is characterized by its resource cost vector $\mathbf{c}_n=[c_{1,n},c_{2,n},\dots,c_{M,n}]^\mathrm{T}$. At any time instance, the active slice set can be represented by a $N$-dimensional vector $\mathbf{s}=[s_1,s_2,\dots,s_N]^\mathrm{T}$, where $s_n$ denotes the number of active slices of type $n$. Correspondingly, the resource assignment can be described as
	\begin{equation}\label{equ:resource_assignment}
	\mathbf{a}\triangleq[a_1,a_2,\dots,a_M]^\mathrm{T}=\mathbf{C}\times\mathbf{s},
	\end{equation}
	where $\mathbf{C}=[\mathbf{c}_1,\mathbf{c}_2,\dots,\mathbf{c}_N]$. Thus, the \textit{space of resource feasibility} is given by
	\begin{equation}
	\mathcal{S}=\{\mathbf{s}:r_m-a_m\ge 0,\quad\forall 1\le m\le M\},
	\end{equation}
	\change{which is illustrated bottom-left in Fig. \ref{fig:system_model}.}
	
		\change{It should be noticed that this is a highly abstracted definition of resource for keeping the generality. In practice, network slicing can be applied both on physical resources, i.e. radio/infrastructure resources~\cite{vural2018dynamic}, and on virtualized resource blocks, i.e. computational capacity~\cite{ha2017end}. The practical design of resource pool, therefore, depends on the use case specification. Generally, all virtualized resource blocks on the same server or server cluster, no matter exploited by which virtual network function (VNF), can be considered as homogeneous and therefore modeled with one dimension of the resource vector $\mathbf{r}$. In contrast, heterogeneous physical resources such as frequency bands and transmission power, must be distinguished with different orthogonal dimensions in $\mathbf{r}$.}
		
		\change{It also worths to note, that the linear resource assignment (\ref{equ:resource_assignment}) formally excludes any resource multiplexing over different slices, which is, especially for physical resources, not only common in practice but also essential for realizing slice elasticity. Nevertheless, as derived in \cite{bega2017optimising}, in the context of inter-slice resource management, an elastic slice that shares resources with other homogeneous slices is equivalent in resource consumption to an inelastic slice with downscaled utility efficiency. Therefore, in this work we consider only inelastic slices for simplification, as elastic slices can be modeled as their inelastic equivalents.}
	
	\subsection{Resource Assignment and Release}\label{subsec:resrc_ass_rel}
	In \change{SlaaS}, resource requests for implementing slices of different types randomly arrive from various tenants. Here we consider a time-frame-based processing of tenant requests: in every time frame, a random number of requests for every type of slices are proposed by tenants.  In this paper, we use the term \textit{operations period} to denote the length of this time frame. Once a request for slice type $n$ arrives, the MNO checks if its idle resources can support it to create such a slice. If not, the request will be immediately declined. Otherwise, the MNO can decide if to accept the request or to decline it. Upon acceptance, the MNO creates a new slice of type $n$, and allocates a corresponding resource bundle $\mathbf{c}_n$ from its idle resource pool to the new slice. 
	
	\change{If the request is declined, no slice will be created. This implies that some tenant may fail to immediately obtain the required network resource for their service, especially when the resource pool is highly occupied. To solve this problem, a mechanism of delayed service upon request decline is required instead of a simple denial. For example, a random-access-alike protocol can be designed to let the tenant resubmit its declined slice creation request after a random delay. Alternatively, the MNO can buffer all declined requests in a waiting queue for future decision. In case a bidding mechanism is valid as proposed in \cite{caballero2017network}, the declined tenant can also reattempt with a raised bid for a better opportunity of acceptance. With such approaches, the binary decision mechanism is able to eventually accept every request after some delay.  As the scope of this paper focuses mainly on the optimization of binary decision, we consider the random delay approach which keeps the request arrivals Poisson distributed, and do not discuss its impact on the latency of slice creation, which worths further studies in future.}
	
	Depending on the business mode, the termination of a slice can either be planned in the request, or randomly happen upon cancellation by the tenant. In this work, we consider the latter case, where every slice of type $n$ has a random lifetime. When a slice of type $n$ is terminated, its resource bundle $\mathbf{c}_n$ will be released and returned to the MNO's idle resource pool.	

	\change{To simplify the analysis and simulation, in this work we assume that all requests for slice termination only arrive and get handled at the beginning/end of operation periods, while requests for slice termination can arrive any time and will be responded by the MNO immediately. Fig. \ref{fig:system_model} briefly concludes this procedure of releasing and assigning resources.}
	
	\subsection{Space of Free Decision and Slicing Strategy}\label{subsec:space_of_free_decision}
	So far, we can describe the MNO's decision on an incoming resource request of slice type $n$ with a binary variable $d\in\{0,1\}$, where $d=0$ denotes decline and $d=1$ stands for acceptance. Upon the decision, the active slice set is updated from its previous value $\mathbf{s}$ to
	\begin{equation}
	\begin{split}
	&g(\mathbf{s},n,d)
	\\=&\begin{cases}
	\mathbf{s}&d=0,\\
	[s_1,\dots,s_{n-1},s_n+1,s_{n+1}\dots,s_N]&d=1.
	\end{cases}
	\label{equ:state-decision-mapping}
	\end{split}
	\end{equation}
	Given a certain space of resource feasibility $\mathcal{S}$, if the decision $d$ is a function of the current active slice set $\mathbf{s}$ and the incoming request $n$, we say that the MNO has a consistent \textit{slicing strategy}
	\begin{equation}\label{equ:strategy}
	d(\mathbf{s},n):\mathcal{S}\times\{1,2,\dots,N\}\rightarrow\{0,1\}.
	\end{equation}
	In this case, $g(\mathbf{s},n,d)=g(\mathbf{s},n)$, i.e. the new active slice set is uniquely determined by the current active slice set and the incoming request.
	
	As the incoming request $n$ is independent of the current active slice set $\mathbf{s}$, the amount of all different possible constructions of the mapping \change{described by Eq.} (\ref{equ:state-decision-mapping}) is $2^{\Vert\mathcal{S}\Vert\times N}$\change{, where $\Vert\mathcal{S}\Vert$ is the number of all $\mathbf{s}\in\mathcal{S}$}. However, as discussed earlier in Sec.\ref{subsec:resrc_ass_rel}, the MNO cannot accept but only decline the request if its idle resources are not sufficient. Hence, we can further restrict the domain of slicing strategy $d$ to the \textit{space of free decision}:
	\begin{equation}
	\mathcal{D}=\{(\mathbf{s},n):\forall\mathbf{s}\in\mathcal{S}, \forall1\le n\le N, d(\mathbf{s},n)\in\mathcal{S}\},
	\end{equation}
	whose size $\Vert\mathcal{D}\Vert$ is slightly smaller than $\Vert\mathcal{S}\Vert\times N$.
	
	\subsection{Utility Model and Long-Term Strategy Optimization}
	Depending on the slice type $n$, every active slice generates a certain utility in every operations period, which we denote with $u_n$.
	\change{
	Depending on the use case, the utility can be flexibly defined in the tenant's point of view as a function of some specified key performance indicator (KPI) such as the network throughput, the average latency or the network reliability. Alternatively, it can also be defined in the MNO's point of view as a direct payoff such as the payment for renting the network resource bundle.
	The} overall utility generated by all slices in an arbitrary operations period $t$ is
	\begin{equation}\label{equ:network_utility}
	u_\Sigma(t)=\sum\limits_{n=1}^Ns_n(t)\cdot u_n.
	\end{equation}
	In this work, our interests focus on selecting the optimal slicing strategy that maximize the expected average overall utility over a long term of $T$ operations periods:
	\begin{equation}\label{equ:fitness}
	d_\text{opt}=\arg\max\limits_{d}\mathrm{E}\left\{\frac{1}{T}\sum_{t=1}^Tu_\Sigma(t)\right\}.
	\end{equation}
\change{Such a strategy is supposed to optimize, depending on the selection of utility function, either the overall performance of the entire sliced network, the economic revenue of the MNO, or other target metric.}
This is a non-convex optimization problem, where no analytic solution is available and heuristic techniques are therefore needed.
	
	\section{Existing Methods and Limits}\label{sec:review}
	\change{
	SlaaS shall be considered as a specific form of cloud computing. The problem of network resource management also commonly exists in the classical cloud environments, including Infrastructure as a Service (IaaS), Platform as a Service (PaaS) and Software as a Service (SaaS). Since over a decade, various approaches have been proposed to schedule and allocate physical and logical resources over different cloud clients\cite{manvi2014resource, jennings2015resource}.}

	\change{	
	Nevertheless, the ubiquitous  features of 5G network slicing are challenging the deployment of classical cloud resource allocation schemes in SlaaS. First, it usually considers almost homogeneous instances and simple resource pool in classical public cloud environments, which simplifies the resource constraints to one-dimensional \cite{inomata2011proposal,he2011real}, or two-dimensional\cite{tomita2011congestion}. In 5G networks, as indicated in\cite{han2017modeling}, a large number of slice types $N$ can be required to support highly heterogeneous mobile services, and the dimension of resource pool $M$ can also be considerably large. These can lead to a high computational complexity of global optimizing algorithms with cascaded loops such as \cite{he2011real}, and reduce their feasibilities. Second, depending on the use scenario, different slices in 5G networks can even have highly heterogeneous constructions of the utility function $u_n$. For instance, the energy efficiency is a critical term in the utility function of mMTC slices, the delay is more important for ultra low-latency reliable communications (URLLC) slices, while the MBB slices are more evaluated regarding the throughput. Classical cloud resource allocating approaches that mostly consider one or few homogeneous cost functions, such as power\cite{chabarek2008power}, throughput\cite{mei2010performance} or resource utilization rate\cite{tomita2011congestion}, for all instances, can be hardly applied in 5G SlaaS. Novel methods are therefore called for, which are expected to be flexible with various constructions of resource constraints and heterogeneous utility functions.
	}
	\change{Recently,} two numerical algorithms have been proposed in \cite{bega2017optimising} to obtain the global optimum of inter-slice resource management
	: the Value Iteration which is an iterative full-search approach, and the Q-Learning which is a model-free online machine learning algorithm. Compared to the Value Iteration, the Q-Learning approach is \change{not only capable to support a flexible selection of optimization target, i.e. the utility function, but also }proven to effectively reduce the computational cost while approximating the optimal performance.
	
	However, the Q-Learning approach has a drawback intrinsically rooting in its action-based optimization framework, that it intends to maximize the average reward that the MNO receives from every decision it makes. This ``decision reward'' lacks of intuitiveness in the business view, and is difficult to map onto common business metrics such as the overall network utility defined in Eq. (\ref{equ:network_utility}).
	
	Furthermore, the Q-Learning algorithm is limited in scalability. The method needs to keep a value table for all possible ``actions'' that the system can take, and to update the values online through an exploration-exploitation process. In the exploration-explotation process, the algorithm has a chance of \change{$\delta$} to intentionally make a wrong or unevaluated decision, so that it guarantees to traverse all possible actions in long-term. By modifying the value of \change{$\delta$}, the algorithm takes its preference between the converging speed and the exploration efficiency. In our case, an action refers to a combination of an arbitrary state in the space of free decision and an arbitrary binary decision, so the size of value table is $2\Vert\mathcal{D}\Vert$. When $\Vert\mathcal{D}\Vert$ grows to a large number, despite of the exploration-exploitation process, the essential time for convergence increases linearly \cite{azar2011speedy} -- which reduces the applicability of the algorithm in practice. This problem can become even worse, when the environment, i.e. the statistical behavior of request arrivals and / or slice terminations, is non-stationary.
	
	\section{Genetic Algorithm}\label{sec:ga}
	Since the 1980s, a category of evolutionary hill-climbing algorithms, known as genetic algorithms (GAs), have been widely applied on various search and optimization problems in the fields of engineering and operations research\cite{goldberg1989genetic}. They have been proved to be efficient in addressing some difficult challenges in such problems, including large state spaces, incomplete state information and non-stationary environments \cite{moriarty1999evolutionary}. 
	\Figure[htbp!](topskip=0pt, botskip=0pt, midskip=0pt)[width=.65\textwidth]{Reproduction.eps}
	{The reproduction procedure in standard GAs\label{fig:standard_ga_reproduction}}
	\Figure[htbp!](topskip=0pt, botskip=0pt, midskip=0pt)[width=.65\textwidth]{Crossover_Mutation.eps}
	{The crossover (left) and mutation (right) procedures in standard GAs\label{fig:standard_ga_crossover_mutation}}

	A GA is intrinsically integrated with a specified encoder, which maps every candidate strategy to an individual binary sequence (code) of a certain length. At the initialization step, a random set of sequences are selected from the codebook, corresponding to the so-called initial \textit{population}. Each candidate strategy is evaluated to obtain its \textit{fitness}, i.e. the value of objective function to optimize. Subsequently, according to the fitness values of the \change{current} population, new populations are iteratively generated. In a standard GA \cite{goldberg1989genetic}, every iteration consists of three sequential steps:
	\begin{enumerate}
		\item \textit{Reproduction}: in this step, every individual strategy in the last population is copied into a new set according to its fitness. The number of copies occurring in the reproduced set is proportional to the fitness value of origin in the last population. The reproduced set has the same size as the last population -- so that the better candidates proliferate \change{through} the reproduction, while the worst outperformed candidates are eliminated. The procedure is briefly illustrated in Fig. \ref{fig:standard_ga_reproduction}.
		\item \textit{Crossover}: in this step, all sequences in the reproduced set are randomly paired. Each pair has a chance to randomly swap a subsequence with each other. By doing so, new sequences are randomly generated, where each ``child'' has a chance to inherit and combine advanced ``genes'' from its both ``parents''. A larger chance of swap (\textit{crossover rate}) leads to a faster convergence to the optimum, while also increasing the risk of premature convergence to local maximums. The procedure is shown in the left part of Fig. \ref{fig:standard_ga_crossover_mutation}.
		\item \textit{Mutation}: where every candidate sequence has a chance to invert one or several random bits of it, which encourages an exploration in the codebook. An increase in \change{either the number of mutation rounds \change{$\beta$} or the chance of one-bit-mutation per round $\gamma$} leads to a reduced risk of premature convergence, while also aggravating the meandering during the convergence -- and therefore raising the risk of drifting away from the global optimum. The procedure is shown in the right part of Fig. \ref{fig:standard_ga_crossover_mutation}.
	\end{enumerate}

	\change{By iterating these steps, a GA is able to approach to the optimum through a winding process. It shall be noted that, differing from classical optimization techniques, GAs do not guarantee to converge at the global optimum due to an endogenous risk of premature convergence. Nevertheless, such risk can be minimized with a variety of techniques. To the readers with further interest on the the converging performance of GAs, we recommend the empirical and analytical studies in \cite{goldberg1989genetic} and \cite{rudolph1994convergence}, respectively.}

	Similar to the Q-Learning algorithm, GAs also possess the advantages of model-free and can be applied online. But differing from most reinforcement learning techniques including the Q-Learning, GAs rely on the quantized ``fitness'' values of different overall strategies instead of the reward value of every single action. This is sometimes considered as a drawback of GAs, because in some applications the fitness function can be difficult to appropriately select \cite{rieser2011comparison}. \change{Nevertheless, this is hardly a flaw in the context of SlaaS, because business metrics such as the long-term average network utility defined by Eq. (\ref{equ:fitness}) are available as fitness functions. On the contrary, it even benefits the deployment of heterogeneous slices to customize the fitness value with different utility functions for various slice types, as discussed earlier in Sec. \ref{sec:review}.}
	
	\change{Another common complain about GAs is that the strategy encoder can be challenging to design. However, in our case of SlaaS here, the binary nature of slicing strategy $d(\mathbf{s},n)$ enables a simple and effective encoder design, which is an important and essential novelty of our work in comparison to existing applications of GA on resource allocation, as we will discuss in the next section.}
	
	\section{Proposed Method}\label{sec:method}
	\subsection{Slicing Strategies as Binary Sequence Codes}
	\change{This is not the first attempt to deploy GA for optimization of resource allocation. Mature solutions have been proposed for allocation of generic resource\cite{lee2003heuristic,liu2005optimization}, radio resource\cite{da2014genetic} and cloud resource\cite{arianyan2012efficient,hachicha2017genetic}. All these approaches consider the problem of global resource optimization, where the system allocates resource blocks from a certain pool to a known set of targets (activities, links, users, etc.). Therefore, they generally aim to optimize the static resource schedule, which is a sequence of resource-target pairs, so every code of theirs represents an individual schedule. In SlaaS, as discussed in Sec. \ref{sec:introduction}, the MNO does not jointly rescale existing slices but make binary decisions to every arriving request for a new slice of random type. The target of optimization here is the slicing strategy $d(\mathbf{s},n)$ as defined in Eq. (\ref{equ:strategy}), and the classical encoding scheme in literature therefore does not apply.}
	
	\change{Noticing from Eq. (\ref{equ:state-decision-mapping}) that $d(\mathbf{s},n)$ is a binary function over a limited domain, every individual slicing strategy can be simply encoded into a binary sequence of length $\Vert\mathcal{S}\Vert\times N$, where each bit represents the MNO's decision to a request for new slice of specific type $1\le n\le N$ with a given active slice set $\mathbf{s}\in\mathcal{S}$. Furthermore, as} 
	discussed earlier in Sec. \ref{subsec:space_of_free_decision}, the MNO can only make a free binary decision when its current active slice set falls in the space of free decision \change{$\mathcal{D}$}. In any other case, the MNO has to decline all incoming requests for new slice creation. \change{Thus, the set of all \emph{feasible} slicing strategies can be enumerated with a codebook of $\Vert\mathcal{D}\Vert$-bit-long binary sequences. }
	
	Therefore, as the first step to encode slicing strategies, we computed the MNO's space of free decision $\mathcal{D}$, which is a limited enumerable set. Subsequently, we mapped $\mathcal{D}$ to the integer set $\left[0, \Vert\mathcal{D}\Vert-1\right]$, \change{which represents the bit positions of a codeword,} as illustrated in Fig. \ref{fig:codec}. \change{By always declining in states outside the space of free decision, i.e. not indexed in the codebook, it guarantees that no decision will break the resource feasibility as far as both the overall resource pool and the list of slice types remain consistent. In case that either of them variates, e.g. when the network infrastructure is maintained or upgraded, both $\Vert\mathcal{S}\Vert$ and $\Vert\mathcal{D}\Vert$ have to be recalculated, and the codebook must be correspondingly updated as well.} 
	\Figure[htbp!](topskip=0pt, botskip=0pt, midskip=0pt)[width=.48\textwidth]{Codec.eps}
	{\change{Every feasible slicing strategy can be uniquely encoded into a binary sequence by a look-up table, where `T' stands for True (accept) and `F' for False (decline). The dark states do not map into the codebook, as they are not in the known space of free decision and therefore always map to `F'.}\label{fig:codec}}
	
	
	\subsection{Genetic Slicing Strategy Optimizer}
	With the code designed above, we implemented a slicing strategy optimizer based on the standard genetic algorithm, \change{which runs in a online mode as Fig. \ref{fig:strategy_opt} illustrates}.
	
	\Figure[htbp!](topskip=0pt, botskip=0pt, midskip=0pt)[width=.9\textwidth]{Strategy_Optimizer.eps}
	{\change{Diagram of the proposed genetic slicing strategy optimizer. Both $d^j$ and $\mathcal{P}^j$ are randomly initialized at $j=1$.}\label{fig:strategy_opt}}
	
	\subsubsection{Initialization} An initial population of candidate strategies, $\mathcal{P}^1$ with certain size $P$, are randomly selected from the pre-generated codebook and kept by the MNO in background for ``\textit{virtual}'' operation. Meanwhile, an initial strategy \change{$d^1$} is randomly generated and applied by the MNO for \textit{actual} operation.
	\subsubsection{Online Fitness Evaluation}
	The MNO sets \change{an} evolution term $T>1$ (normalized to one operations period), and records its active slice set at the beginning of every evolution term. As the network runs, the MNO responses every incoming tenant request according to its currently applied slicing strategy, and meanwhile makes an individual``virtual'' decision in the background according to every candidate strategy in the current population. For every single candidate strategy, the MNO tracks the simulated utility every operations period. At the end of the $j^\text{th}$ evolution term,
	\change{letting $\mathcal{P}^j$ denote the current population, the optimizer evaluates every strategy candidate $p_i^j\in\mathcal{P}^j$ with the average utility it generated (or simulated) over the last evolution term, i.e. the last $T$ operations period:
	\begin{equation}
		\bar{u}_i^j=\frac{1}{T}\sum\limits_{t=(j-1)T+1}^{jT}u_{\Sigma,p_i^j}(t),\quad \forall 1\le i\le P, j\in\mathbb{N^+},
	\end{equation}
	which is taken as the \emph{fitness} value (see Fig. \ref{fig:standard_ga_reproduction}). This implies that, the more utility efficient a candidate strategy performed in the $i^\text{th}$ evolution term, the higher fitness value it gets.}

	\subsubsection{Evolution}
	First, the best candidate in $\mathcal{P}^j$ with respect to the fitness is selected to update the strategy for \textit{actual} operation in the next evolution term:
	\begin{equation}
	\change{d^{j+1}}=\arg\max\limits_{p_i^j\in\mathcal{P}^j}\bar{u}_i^j.
	\end{equation}
	\change{That is, the candidate strategy that had generated / simulated the most utility in the $j^\text{th}$ evolution term is applied by the MNO for its actual operation in the $(j+1)^\text{th}$ evolution term.}
	
	Afterwards, a reproduction $\tilde{\mathcal{P}}^{j}$ of $\mathcal{P}^j$ is generated with respect to the normalized fitness values. The \change{reproduction} numbers of an arbirary $p_i^j\in\mathcal{P}^j$ in $\tilde{\mathcal{P}}^j$ is
	\begin{equation}
	A_i^j=\text{round}\left\{P\times\frac{\bar{u}_i^j+\epsilon}{\sum_{i=1}^{P}\bar{u}_i^j+P\change{\times}\epsilon}\right\},
	\end{equation}
	where $\epsilon$ is a small number to mitigate error in the rare case that $\sum_{i=1}^{P}\bar{u}_i^j=0$. \change{The rounding operation here intends to ensure the number of copies $A_i^j$ to be an integer.}  \change{As shown in Figs.\ref{fig:standard_ga_reproduction}--\ref{fig:standard_ga_crossover_mutation}, the} elements in $\tilde{\mathcal{P}}^{j}$ are then shuffled and paired, each pair has a chance of $\alpha$ to execute the crossover operation. After the crossover, every candidate strategy in the new population experiences $\beta$ turns of mutation, in each turn the candidate strategy has an independent chance of $\gamma$ to invert one random bit of it\change{, as shown in Fig. \ref{fig:standard_ga_crossover_mutation}}. The resulted set of strategies is taken to update the population $\mathcal{P}^{j+1}$ for \textit{virtual} operation in the next evolution term.
	
	\section{Performance Evaluation}\label{sec:evaluation}
	\subsection{Setup of Simulation Environment}
	Towards a brief and convincing demonstration with minimized computational complexity, we considered a MNO with one-dimensional normalized resource pool:
	\begin{equation}
	\mathbf{r}=[r_1]=[1], 
	\end{equation}
	which accepts two different slice types, i.e. $M=1$ and $N=2$. Thus, the resource cost vector of each slice type $\mathbf{c}_n$ is also one-dimensional, which we set to $\mathbf{c}_1=\mathbf{c}_2=[0.3]$. A small space of resource feasibility $\mathcal{S}$ with size of 10 and a small space of free decision $\mathcal{D}$ with size of 12 can be then obtained, as listed in Tab.\ref{tab:spaces}. Under this specification, the number of all feasible slicing strategies $d$ sums to $2^{12}=4096$. 
	\begin{table}[!htpb]
		\centering
		\begin{tabular}{c|c}
			\toprule[2px]
			Elements in $\mathcal{S}$ & Elements in $\mathcal{D}$\\\hline
			\multirow{3}{.2\textwidth}{[0,0], [0,1], [0,2], [0,3], [1,0], [1,1], [1,2], [2,0], [2,1], [3,0]}&\multirow{3}{.25\textwidth}{[0,0,1], [0,0,2], [0,1,1], [0,1,2], [0,2,1], [0,2,2], [1,0,1], [1,0,2], [1,1,1], [1,1,2], [2,0,1], [2,0,2]}\\&\\&\\
			\bottomrule[2px]
		\end{tabular}
		\caption{The spaces of resource feasibility and free decision under the simulation specification}\label{tab:spaces}
	\end{table}
	
	We considered the periodical utilities of the two slice types as $u_1=2$ and $u_2=1$, respectively, so that the slice type $1$ is twice so utility efficient as the slice type $2$. Furthermore, we set the length of an evolution term to $T=6$ operations periods.
		
	\subsection{Definition of Service Scenarios}
	Similar to \cite{bega2017optimising}, we assumed the arrivals of requests for slice creation as Poisson processes, i.e. for every slice type $n\in\{1,2\}$, the number of arriving requests $k_n$ over one operations period is Poisson distributed:
	\begin{equation}
	P(k_n\text{ requests arrive}) = e^{-\lambda_n}\frac{\lambda_n^{k_n}}{k_n!},\quad\forall 1\le n\le N.
	\end{equation}
	Meanwhile, we assumed every slice of arbitrary type $n\in\{1,2\}$ to have a random lifetime (normalized to one operations period) that obeys the exponential distribution:
	\begin{equation}
	f(\tau_n=t_n)=\frac{1}{\mu_n} e^{-\frac{t_n}{\mu_n}},\quad\forall t_n\in\mathbb{N}^+,\forall 1\le n\le N.
	\end{equation}
	For our simulations, we defined three service scenarios with different parameter sets $[\lambda_1,\lambda_2,\mu_1,\mu_2]$, as listed in Tab. \ref{tab:scenarios}. 
	\begin{table}[!htpb]
		\centering
		\begin{tabular}{c|c|c|c|c}
			\toprule[2px]
			Scenario & $\mathbf\lambda_1$ & $\lambda_2$ & $\mu_1$ & $\mu_2$ \\\hline
			\#1&0.5&2&2&10\\
			\#2&0.3&1&2&3\\
			\#3&1&0&2&5\\
			\bottomrule[2px]
		\end{tabular}
		\caption{The model parameters of slice request arrivals and slice terminations in different scenarios}\label{tab:scenarios}
	\end{table}
	
	\subsection{Effectiveness}\label{subsec:effectiveness}
	To demonstrate the effectiveness of our proposed method, we simulated two genetic slicing strategy optimizers in scenario \#1: one with $10$ candidate strategies in every generation, and the other with a larger population size of $50$. Each optimizer was initiated with a random population of candidate strategies and a fully idle resource pool, then evolved 20 generations. Aiming at a fast convergence, both genetic optimizers were set to have full crossover rate $\alpha=1$ and $\beta=1$ round of mutation with rate of $\gamma=0.1$. We repeated this simulation 500 times for Monte-Carlo test, and tracked the long-term average network utility defined in Eq. (\ref{equ:network_utility}). Meanwhile, as a benchmark, the global optimum out of all 4096 feasible strategies was obtained by full-search through 500 times of the same Monte-Carlo test. Besides, we also tested three ``naive'' reference strategies as baselines for performance comparison:
	\begin{itemize}
		\item \textit{Greedy}: accepting all incoming requests, so long as the resource pool supports
		\item \textit{Conservative}: accepting all requests for type 2 slices, while declining all requests for type 1 slices
		\item \textit{Opportunistic}: accepting all requests for type 1 slices, while declining all requests for type 2 slices
	\end{itemize}
	\change{As the benchmark strategies do not evolve, they remain constant over all generations.} 
	The results are illustrated in Fig. \ref{fig:effectiveness}. It can be observed that both genetic optimizers started on poor initial utility levels, but then converged to competitive slicing strategies with satisfying performances quickly (within 4 generations). The genetic optimizers failed to achieve the global optimum of utility efficiency within the simulated progress, converging to a local maximum. Nevertheless, from the fourth generation of evolution on, i.e. after evaluating 30 or 150 out of the 4096 strategies, both genetic optimizer outperformed all three static naive reference strategies with long-term average network utilities over 90\% with respect to the global optimum. Additionally, comparing the two optimizers with each other, it can be observed that the increase in population size boosts the convergence.
	\Figure[htbp!](topskip=0pt, botskip=0pt, midskip=0pt)[width=.48\textwidth]{demonstration_500mc_with_markers.eps}
	{\change{Network utilities generated by proposed genetic optimizer in comparison to those under reference slicing strategies\label{fig:effectiveness}}}
	
	\subsection{Evolution of the Entire Population}\label{subsec:exploration}
	An important feature of GA is that not only the best candidate but also the entire population evolve in every iteration. Fig. \ref{fig:exploration} shows the performance distribution of the genetic optimizer's population with 50 strategies in different generations. A significant approach towards an overall ``good'' strategy set can be observed. This phenomenon reveals a potential of our genetic optimizer in generating training sets for initialization and updating, when it is jointly applied with other machine learning methods.
	\Figure[htbp!](topskip=0pt, botskip=0pt, midskip=0pt)[width=.48\textwidth]{exploration.eps}
	{As the genetic optimizer runs, the entire population generally approach towards a ``good'' strategy set.\label{fig:exploration}}
	
	\subsection{Robustness against Non-Stationarity}
	As mentioned earlier in Sec. \ref{sec:ga}, GAs are known to be robust against non-stationary environments. In the context of slicing strategy optimization, this refers to time-varying statistical behavior of resource requests and slice terminations. To test our genetic slicing strategy optimizer under such conditions, we conducted a simulation over 60 generations of evolution, i.e. 360 operations periods. For the first 20 generations, the scenario was set to \#1, so that the same global optimal strategy obtained in Sec. \ref{subsec:effectiveness} remained valid; during the generations 21 to 40, the scenario was set to \#2; during the generations 41 to 60, the scenario was updated again to \#3. We deployed a genetic optimizer in this scenario, which was specified to the parameters $[\alpha,\beta,\gamma]=[1,1,0.1]$ and a population size of 50. Its performance was compared with those of the global optimal strategy obtained in scenario \#1, as well as of the three aforementioned naive reference strategies. The results given by 500 times  of Monte-Carlo tests are illustrated in Fig. \ref{fig:non-stationary}. It can be observed that the genetic optimizer succeeded to quickly adapt with environmental variations, and hence remained on a high performance level. In contrast, the scenario-specified optimum gave a poor dynamic performance when the environment changed. Similarly, the performances of all static reference strategies also turned out to strongly rely on the environment. 
	\Figure[htbp!](topskip=0pt, botskip=0pt, midskip=0pt)[width=.48\textwidth]{non_stationarity_with_markers.eps}
	{\change{Genetic optimizers are feasible in non-stationary scenarios, outperforming all static reference strategies.\label{fig:non-stationary}}}


	\subsection{Scalability and Enhancements}
	To test the computational scalability of our genetic slicing strategy optimizer, we set a complexer environment with significantly smaller slice scales $\mathbf{c}_1=\mathbf{c}_2=0.03$. Under this specification, the MNO has a space of resource feasibility $\mathcal{S}$ with size 595, a space of free decision $\mathcal{D}$ with size 1122. The amount of its possible slicing strategies $d$ therefore sums to an astronomical figure of $2^{1122}$. We also correspondingly scaled the utility efficiencies to $[u_1,u_2]=[0.2,0.1]$, and set the service scenario parameters to $[\lambda_1,\lambda_2,\mu_1,\mu_2]=[2.5,10,2,10]$. 
	
	Then we tested two genetic optimizers with population sizes of 10 and 50, respectively. Both optimizers were set to $[\alpha,\beta,\gamma]=[1,1,0.1]$. Once again, we took the reference strategies ``Greedy'', ``Conservative'' and ``Opportunistic'' as benchmarks. No global optimum was evaluated, as the computational cost of full-search for the optimum is unbearably high. As illustrated in Fig. \ref{fig:scalability_local_convergence}, both optimizers succeeded to quickly converge within 10 generations, but only to reach local maximums that are much worse than all reference strategies.
	    	
	\Figure[htbp!](topskip=0pt, botskip=0pt, midskip=0pt)[width=.48\textwidth]{scalability_local_convergence_with_markers.eps}
	{\change{When the solution space is huge, genetic optimizers are still able to evolve fast, but can easily converge to poor local maximums.\label{fig:scalability_local_convergence}}}
    
    This phenomenon has its origin in the fact, that as the size of strategy space grows, both the amount of local maximums and the average distance between a random strategy and the global optimum increase. As a consequence, the risk of premature convergence also rises. Additionally, as the GA initiates with a random population, it can easily converge to a poor level.
    
    To counter this effect, efforts can be made in two aspects: \begin{enumerate*}
    \item to improve the initial population, and
    \item to mitigate early convergences at local maximums.
    \end{enumerate*}  The first one can be achieved by involving one or several reference strategies into the initial population, so that the performance evolves from the benchmark level. For the second, either a lower crossover rate or a higher mutation rate can help. Additionally, it is a common technique in GAs to preserve one or several ``elite'' individuals in every generation from the crossover and mutation operations, and directly put it into the next generation, in order to suppress the random degradation that may caused by mutations\cite{deb2002fast}.
    
    So we repeated the aforementioned simulation, manually involving the reference strategy ``Greedy'' in the initial random populations of both optimizers. Both optimizers were configured to $[\alpha,\beta,\gamma]=[0.9,1,0.1]$ and to preserve one best individual in every generation of population. The results are depicted in Fig. \ref{fig:scalability_improved_convergence}. It can be observed that the genetic optimizers either outperformed the benchmark or at least draw it with these simple enhancements. Both optimizers converged within 6 generations.

	\change{Another phenomenon that worths to notice is that, the genetic optimizer with smaller population may temporarily outperform the one with larger population in the first generations, as Fig. \ref{fig:scalability_improved_convergence} exhibits. This is determined by the stochastic and winding nature of evolving process, and the fact that the initial population of candidate strategies are randomly selected as well. Nevertheless, as we can learn from Figs. \ref{fig:scalability_local_convergence} and \ref{fig:scalability_improved_convergence}, a large population brings a long-term utility gain when the optimizer eventually converges.}

	\Figure[htbp!](topskip=0pt, botskip=0pt, midskip=0pt)[width=.48\textwidth]{scalability_improved_convergence_with_markers.eps}
	{\change{With minor enhancements, genetic optimizers guarantee to outperform any certain strategy. The convergence can be further improved at the expense of population size without significantly increased time complexity. \label{fig:scalability_improved_convergence}}}	
    
    \subsection{Summary}
    So far, we can assert that our genetic optimizer guarantees to converge to outperform any certain static strategy, while the essential time for convergence only slightly increases with the size of solution space. It is also worth to note that we can improve the convergence by extending the population. As the evaluation of different candidate strategies in the same generation can be easily parallelized \cite{goldberg1989genetic}, the upscaling of population impacts little on the time complexity of our proposed method, making it highly scalable and practical for complex realistic applications.

	\section{Conclusion and Outlooks}\label{sec:conclusion}
	In this paper, we have presented a novel online genetic slicing strategy optimizer to maximize the long-term network utility in \change{SlaaS}. The proposed approach has been evaluated through numerical simulations, exhibiting a satisfying approximate to the global optimum, a fast convergence, a timely adaptation to environment variation and a good scalability. \change{It encodes slicing strategies instead of resource schedules into binary sequences, which enables genetic optimization for inter-slice resource management based on tenant requests and MNO's binary decisions.} Besides, it requires no a priori knowledge about the traffic or utility model.
	
	As follow-up work, it remains interesting to enhance the convergence performance of the proposed slicing strategy optimizer with advanced operations and techniques in genetic search, such as fitness scaling, diploid evolution and sequence reordering\cite{goldberg1989genetic}. \change{Especially, it worths an attempt to ameliorate the rate of convergence of GA with heuristic searching as reported in \cite{lee2003heuristic}, in order to meet the real-time requirement of network resource management.}  Besides, as referred in Sec. \ref{subsec:exploration}, there is also a great potential to combine our genetic slicing strategy optimizer with other machine learning approaches such as reinforcement learning and artificial neural networks. \change{Additionally, as mentioned in Sec.\ref{subsec:resrc_ass_rel}, different mechanisms to grant declined tenants slices after delays and their impacts on the business case of SlaaS worth further studies, as well.}

	\bibliographystyle{IEEEtran}
	\bibliography{references}
	
    \vfill

	\begin{IEEEbiography}[{\includegraphics[width=1in,height=1.25in,clip,keepaspectratio]{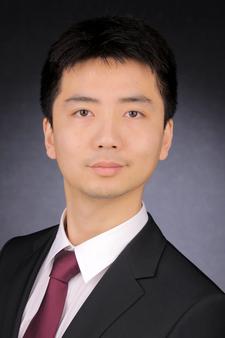}}]{Bin Han} (M'15) received his B.E. degree in Electronic Science and Technology in 2009 from the Shanghai Jiao Tong University, China, and his M.Sc. degree in Electrical and Information Engineering in 2012 from the Technische Universit\"at Darmstadt, Germany. In 2016 he was granted the Dr.-Ing. degree in Electrical and Information Engineering from the Kalsruhe Institute of Technology, Germany. Since July 2016 he joined the Institute of Wireless Communication (WiCon), University of Kaiserslautern, Germany, and has been participating in multiple EU Horizon 2020 research projects for 5G mobile networks. Currently, he is working as senior lecturer at WiCon. His current research interests are in the broad area of wireless communication systems and signal processing, with special focus on inter-slice resource management.
	\end{IEEEbiography}

	\vfill

	\begin{IEEEbiography}[{\includegraphics[width=1in,height=1.25in,clip,keepaspectratio]{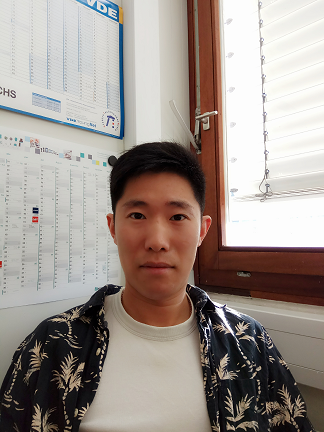}}]{Ji Lianghai} (S'17) received his B.Sc degree from the Shandong University, China in 2010 and his M.Sc. degree from the University of Ulm, Germany in 2012. He is at this moment working for his Ph.D degree at the Institute of Wireless Communication, Technische Universit\"at Kaiserslautern, Germany. He worked for European 5G flagship projects METIS and METIS- 2 and some other 5G projects with industrial partners. He is currently representing University of Kaiserslautern in German 5G project 5G-NetMobil.
	\end{IEEEbiography}
	
    \vfill
	 
	\begin{IEEEbiography}[{\includegraphics[width=1in,height=1.25in,clip,keepaspectratio]{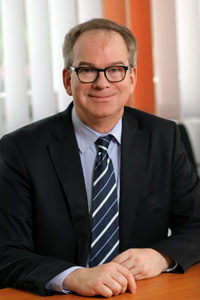}}]{Hans D. Schotten} (S'93--M'97)  received the Diplom and Ph.D. degrees in Electrical Engineering from the Aachen University of Technology RWTH, Germany in 1990 and 1997, respectively. Since August 2007, he has been full professor and head of the Institute of Wireless Communication at the Technische Universit\"at Kaiserslautern. Since 2012, he has additionally been Scientific Director at the German Research Center for Artificial Intelligence heading the ``Intelligent Networks'' department.
	\end{IEEEbiography}
	
	\vfill
	
	\EOD
	
\end{document}